\pdfoutput=1

\documentclass[11pt]{article}

\usepackage[]{ACL2023}

\usepackage{times}
\usepackage{latexsym}
\usepackage{amssymb}

\usepackage[T1]{fontenc}

\usepackage[utf8]{inputenc}

\usepackage{microtype}
\usepackage{svg}
\usepackage{inconsolata}
\usepackage{multirow}
\usepackage{makecell}
\usepackage{amsmath,graphicx}
\usepackage{relsize,etoolbox}
\AtBeginEnvironment{quote}{\smaller}

\newcommand{\nameShort}{\textsc{Sup-NatInst}}
\newcommand{\preNameShort}{\textsc{NatInst}}
\newcommand{\modelName}{\textsc{T$k$-Instruct}}
\usepackage{pifont}

\newcommand{\cmark}{\ding{51}}%
\newcommand{\xmark}{\ding{55}}%

\title{Improving Cross-Task Generalization with Step-by-Step Instructions}

\author{
	\bf Yang Wu$^1$\thanks{~~~This work was conducted during the internship of Yang Wu at Huawei Cloud} ~ Yanyan Zhao$^1$\thanks{~~~Corresponding Author} ~ Zhongyang Li$^2$   ~ Bing Qin$^{1}$ ~ Kai Xiong$^{1,3}$ \\
	$^1$ Harbin Institute of Technology  \quad \quad $^2$  Huawei Cloud\\
        $^3$  Singapore Management University \\
        $^1$ \{\tt ywu, yyzhao, qinb, kxiong\}@ir.hit.edu.cn \\
	$^2$ \tt lizhongyang6@huawei.com \\
}

\begin{document}
\maketitle
\begin{abstract}
Instruction tuning has been shown to be able to improve cross-task generalization of language models. However, it is still challenging for language models to complete the target tasks following the instructions, as the instructions are general and lack intermediate steps. To address this problem, we propose to incorporate the step-by-step instructions to help language models to decompose the tasks, which can provide the detailed and specific procedures for completing the target tasks. The step-by-step instructions are obtained automatically by prompting ChatGPT, which are further combined with the original instructions to tune language models. The extensive experiments on \nameShort{} show that the high-quality step-by-step instructions can improve cross-task generalization across different model sizes. Moreover, the further analysis indicates the importance of the order of steps of the step-by-step instruction for the improvement. To facilitate future research, we release the step-by-step instructions and their human quality evaluation results.
\end{abstract}

\section{Introduction}

How to improve cross-task generalization of language models is a vital but difficult problem, which has attracted more and more attention from the NLP community~\cite{ye-etal-2021-crossfit, mishra2022cross, supernaturalinstructions, sanh2022multitask, chung2022scaling}.~\citet{mishra2022cross} construct the \preNameShort{} dataset consisting of 61 various NLP tasks to evaluate the cross-task generalization of language models, which are trained on a part of tasks and evaluated on other tasks. ~\citet{supernaturalinstructions} extend \preNameShort{} and build a much larger dataset, namely \nameShort{}, which includes 1616 NLP tasks. The studies ~\cite{mishra2022cross, supernaturalinstructions} conducted on \preNameShort{} and \nameShort{} show that instruction tuning can improve the generalization of language models to new tasks.

However, the natural language instructions adopted by previous work~\cite{supernaturalinstructions}, of which the main elements are the task definitions, are general and lack intermediate steps, which makes it challenging for language models to follow the instructions and complete the target tasks. Hence, we propose to incorporate the step-by-step instructions to make the instructions more detailed and specific. The step-by-step instructions can provide the \textit{task-level} intermediate problem-solving steps without depending on any specific example, which are easy to understand and follow. The step-by-step instructions are also significantly different from chain-of-thought~\cite{wei2022chain}, which consists of the \textit{example-level} intermediate reasoning steps. Moreover, the step-by-step instructions are automatically generated by ChatGPT\footnote{https://chat.openai.com}, which is trained to align with instructions by using reinforcement learning from human feedback~\cite{stiennon2020learning}. We treat ChatGPT as a translator and ask it to first understand the intents behind the original instructions and then write the step-by-step instructions for completing the target tasks. To obtain more desirable results, we further progressively refine the step-by-step instructions through multiple interactions with ChatGPT. The final step-by-step instructions are combined with the original instructions to tune language models.

\begin{figure*}[t]
\centering
\includegraphics[width=\linewidth]{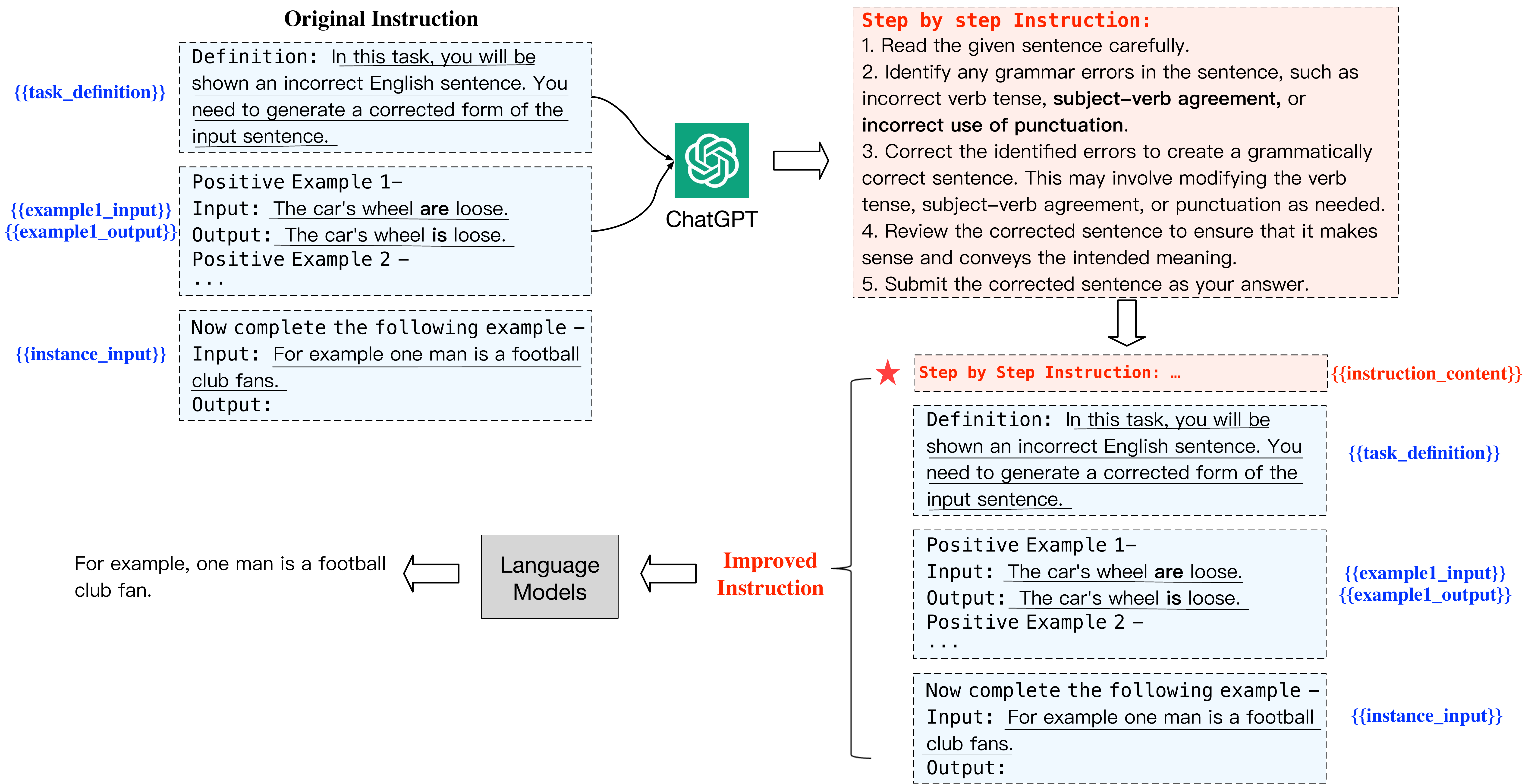} 
\caption{\textit{Step-by-Step Instruction Tuning} consists of two steps: (1) prompt ChatGPT to obtain the step-by-step instruction based on the task definition and the positive examples(\S\ref{sec:obtaining}, \S\ref{sec:refining}); (2) combine the step-by-step instruction with the original instruction to tune the language models(\S\ref{sec:tuning}). The contents of the original instruction elements are marked with \underline{underlines} and we omit the refining process for simplicity. }
\label{begin}
\end{figure*}

To have an intuitive understanding of our approach, namely \textit{Step-by-Step Instruction Tuning}, we show an example in Figure~\ref{begin}. As shown in this figure, the original instruction includes the task definition, positive examples, and instance. However, the task definition is very short and general. Hence, we pass the task definition and the positive examples to ChatGPT to obtain the step-by-step instruction. The obtained step-by-step instruction is detailed and specific, which shows the intermediate steps of completing the task and even points out the possible grammar errors such as subject-verb agreement and incorrect use of punctuation in Step 2. We believe the detailed step-by-step instructions can help language models to complete this task.

We conduct extensive experiments on \nameShort{}~\cite{supernaturalinstructions} to evaluate our proposed approach. The experimental results demonstrate the effectiveness of \textit{Step-by-Step Instruction Tuning}, which improves the cross-task generalization of T5-LM with different model sizes\footnote{We focus on improving the instructions and our approach also could be applied to other language models.}. To comprehensively understand our approach, we analyze various important factors affecting the model performance such as the order of steps. The analysis indicates that shuffling the order of steps leads to a performance drop since it corrupts the correctness of the step-by-step instruction. Besides, we also attempt to leverage ChatGPT to generate positive examples and present the results in Appendix~\ref{generate_example}. 




The main contributions of this paper are described as follows: (1) we are the first to incorporate the step-by-step instructions to improve cross-task generalization and the extensive experiments demonstrate its effectiveness; (2) we are the first to propose to automatically generate and refine the step-by-step instructions through interactions with ChatGPT; (3) we conduct a comprehensive human evaluation to analyze the quality of the step-by-step instructions; (4) we release the step-by-step instructions and the results of the human evaluation for facilitating future research on improving generalization with the step-by-step instructions.

\section{Related Work}

\paragraph{Instruction tuning.} Instruction tuning has shown its effectiveness in improving the generalization of language models. ~\citet{sanh2022multitask} and ~\citet{wei2022finetuned} both collect a large dataset of different tasks and split a part of the tasks as the training set and take the remaining tasks as the test set. They mix the data of the training set and train the language models using multi-task learning. The tuned models are evaluated on the test set to estimate their zero-shot performance. Their experimental results show instruction tuning can improve zero-shot performance of large language models. ~\citet{supernaturalinstructions} also build a meta-dataset, namely SUP-NATINST, which consists of various NLP tasks and they evaluate the few-shot performance of language models given the instruction, but the format of the instruction is different from the previous two works. In contrast to taking the simple manual prompt as the instruction~\cite{sanh2022multitask, wei2022finetuned}, the instruction of SUP-NATINST consists of the task definition, the positive examples, and the negative examples. In this paper, we mainly focus on this format of instruction and propose to improve the cross-task generalization of language models with the step-by-step instructions. Our approach is also significantly different from ~\citet{mishra-etal-2022-reframing}. Because \citet{mishra-etal-2022-reframing} manually reframe the instructions of the evaluation tasks of \preNameShort{}~\cite{mishra2022cross} to make them more suitable for prompting GPT models, which is hard to extend to new tasks. Our approach automatically obtains the step-by-step instructions via prompting ChatGPT with a series of task-agnostic prompts, which can be easily adopted for other tasks.

\paragraph{Chain-of-thought prompting.} Recently, chain-of-thought (CoT) prompting~\cite{wei2022chain} has shown impressive results on many complicated reasoning tasks such as the math word problem, which incorporates a series of manually written intermediate reasoning steps for the demonstration examples to unlock the reasoning ability of large language models. However, the adopted demonstrations are task-specific and carefully designed, which are hard to obtain. In contrast to it, ~\citet{kojima2022large} propose Zero-shot-CoT, which first obtains the intermediate reasoning steps via prompting the large language models and then incorporates such intermediate reasoning steps to get the answer. ~\citet{zhang2022automatic} introduce Auto-CoT to sample questions with diversity and generate reasoning chains to construct demonstrations. 

Even though both CoTs and our approach propose to decompose the task into multiple steps~\cite{zhou2022least, khot2022decomposed}, our approach is fundamentally different from CoTs. Firstly, the step-by-step instruction does not depend on any specific example, which is a general problem-solving procedure for completing the target task, while the chain-of-thought is a series of intermediate reasoning steps for a specific example. Secondly, our approach aims to improve cross-task generalization of language models to unseen tasks, while CoTs are proposed to complete complex reasoning tasks. Moreover, we focus on improving the generalization of smaller language models, while CoTs are mainly applied to large language models of $\sim$100B parameters~\cite{wei2022chain}.



\section{Method}
In this section, we first introduce how to obtain and refine the step-by-step instructions automatically via prompting ChatGPT with a series of task-agnostic prompts (\S\ref{sec:obtaining} and \S\ref{sec:refining}). Then, we conduct a detailed analysis to evaluate the quality of the obtained step-by-step instructions (\S\ref{sec:analysis}). Finally, we propose \textit{Step-by-Step Instruction Tuning} to incorporate such step-by-step instructions to improve cross-task generalization (\S\ref{sec:tuning}). 

\subsection{Step-by-Step Instruction Obtaining} \label{sec:obtaining}
 We carefully design the prompt to ask ChatGPT to generate an easy-to-follow step-by-step instruction for the target task based on the task category\footnote{The content of the task category is most often covered by the task definition. It is used here to induce ChatGPT to attend the task definition of the bottom of the prompt.} and task definition. In this prompt, we add some constraints to help ChatGPT to generate more desirable outputs. For example, we specify that the step-by-step instruction is used for instructing the generative pre-trained language models to prevent ChatGPT from generating unapplicable intermediate steps, such as using the search engines. The full adopted prompt is as follows. 
 


\begin{quote}
    Please provide a step-by-step instruction for completing the \{\{task\_category\}\} task. The generated instruction will be directly used as the input of the generative pre-trained language models. The instruction should be simple and easy to understand, without any specific examples. 
    
    Note that: The instruction should only focus on the current example. Do not contain the step of iterating through the dataset. 

    \{\{task\_category\}\}: \textcolor{blue}{\{\{task\_definition\}\}}
\end{quote}

\{\{task\_category\}\} and \{\{task\_definition\}\} will be replaced with the task category and definition of the specific target task.

\subsection{Step-by-Step Instruction Refining} \label{sec:refining}

Even though ChatGPT is asked to generate appropriate step-by-step instruction, ChatGPT sometimes does not follow the prompt. Hence, we propose to refine the step-by-step instruction though multiple interactions with ChatGPT. Firstly, ChatGPT could instruct to iterate the process through the dataset. To address this problem, we design the prompt to make ChatGPT refine the step-by-step instruction to make sure that it is suitable for a single example.

\begin{quote}
Refine the instruction to make sure the instruction is applicable for a single example and does not instruct to repeat the process through the dataset.
\end{quote}

Secondly, we utilize the positive examples to help ChatGPT to more comprehensively understand the task leading to better step-by-step instruction. Similarly, \{\{example1\_input\}\}, \{\{example1\_output\}\}, \{\{example2\_input\}\} and \{\{example2\_output\}\} will be filled with the contents of the specific examples.

\begin{quote}
Refine your instruction according to the given examples and return the full refined instruction. The refined instruction should be simple and easy to understand, without any specific examples. 

Example 1:

Input: \textcolor{blue}{\{\{example1\_input\}\}}

Output: \textcolor{blue}{\{\{example1\_output\}\}}

Example 2:

Input: \textcolor{blue}{\{\{example2\_input\}\}}

Output: \textcolor{blue}{\{\{example2\_output\}\}}

\end{quote}
Thirdly, to avoid the step-by-step instruction containing any specific example and make it general, we further ask ChatGPT to check and refine the step-by-step instruction.

\begin{quote}
Refine the instruction to make sure the instruction does not contain any specific example.
\end{quote}

Lastly, we fetch the final step-by-step instruction from ChatGPT using the following prompt.

\begin{quote}
Output the refined instruction.
\end{quote}

\begin{table}[tbp]
	\centering
        \resizebox{\linewidth}{!}{
	\begin{tabular}{c|c|c}
		\hline
		\multicolumn{1}{l|}{\multirow{1}{*}{Statistic}} & \multicolumn{1}{r}{Train} & \multicolumn{1}{r}{Test}  \\
		\hline
		\hline
		\multicolumn{1}{l|}{Total number of tasks}	 &             \multicolumn{1}{r}{756} &             \multicolumn{1}{r}{119}  \\ 
		\multicolumn{1}{l|}{Total number of categories}	 &             \multicolumn{1}{r}{60} &             \multicolumn{1}{r}{12}  \\ 
		\multicolumn{1}{l|}{Average word count per definition}	 &             \multicolumn{1}{r}{66.2} &             \multicolumn{1}{r}{65.6}  \\ 
		\multicolumn{1}{l|}{Average word count per step-by-step instruction}	 &             \multicolumn{1}{r}{118.2} &             \multicolumn{1}{r}{91.6}  \\ 
  		\multicolumn{1}{l|}{Average number of steps per step-by-step instruction}	 &             \multicolumn{1}{r}{6.0} &             \multicolumn{1}{r}{5.4}  \\ 

		 \hline
        \end{tabular}
        }
  	\caption{Statistics of SUP-NATINST (English track).  }
	\label{tab:statistics}
\end{table}








\begin{figure}[t]
\centering
\includegraphics[width=\linewidth]{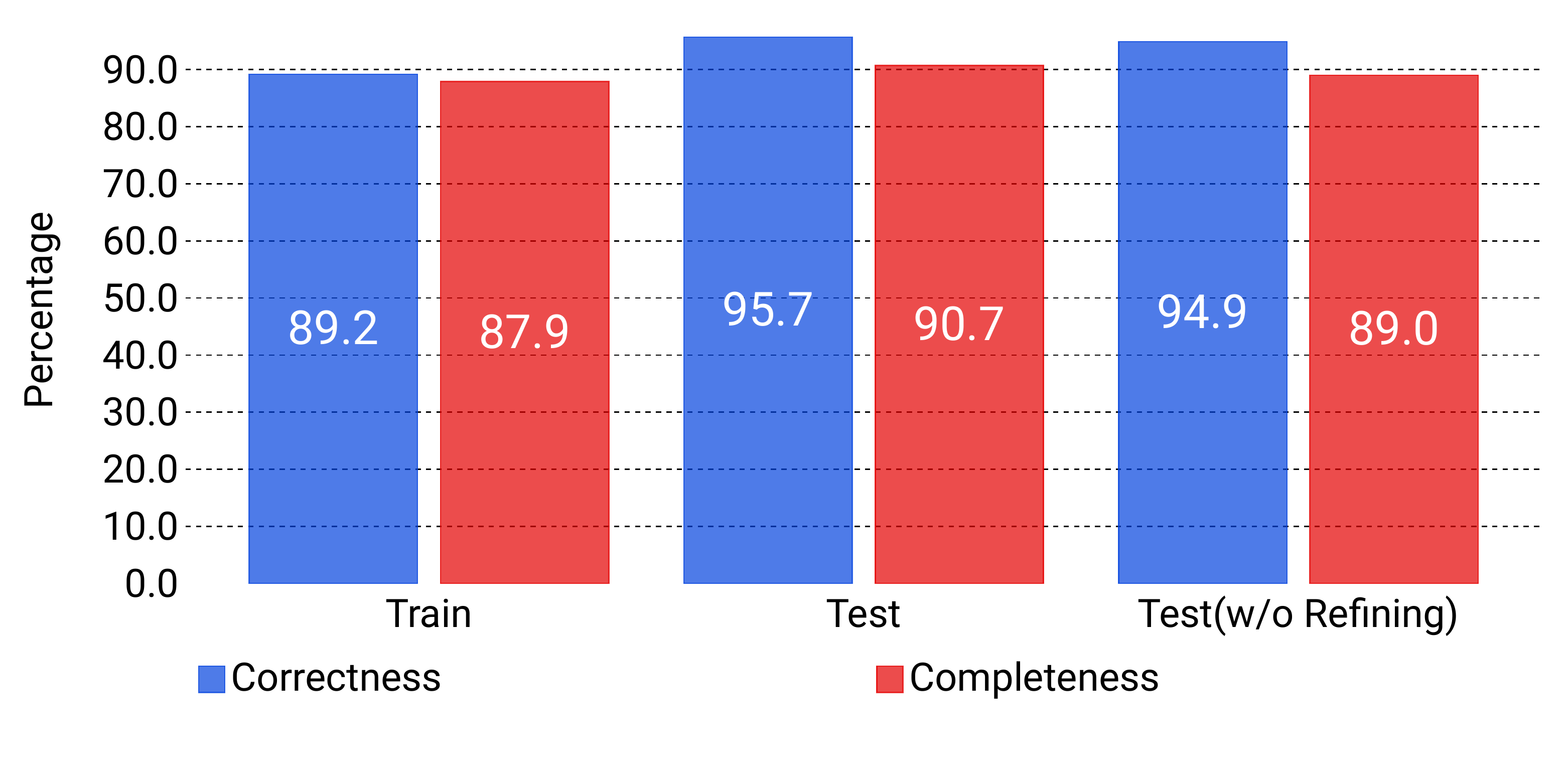} 
\caption{Human evaluation of the step-by-step instructions.  }
\label{instruction_quality}
\end{figure}

\subsection{Analysis of Step-by-Step Instruction} \label{sec:analysis}
We conduct an in-depth analysis of the obtained step-by-step instructions including statistical analysis and human evaluation. Table~\ref{tab:statistics} shows the results of the statistical analysis. The total number of categories means the number of task types in the dataset, which can be keyword tagging and question rewriting. As for the task category and task, there is no overlap between training set and test set. 

To quantitatively evaluate the quality of the step-by-step instructions, we adopt correctness and completeness as the metrics. Specifically, if both each step of the step-by-step instruction and the order of the steps are right for completing the target task, we consider that it is correct. And the completeness metric indicates whether the step-by-step instruction contains all the necessary information of the original instruction such as the constraints. The human evaluation is conducted by three well-educated annotators and the results are presented in Figure~\ref{instruction_quality}. More details about the annotation process are presented in Appendix~\ref{annotation}. We sample 60, 20, and 20 examples from the training set, test set, and test set (w/o Refining) as a shared annotation part of all annotators and split the remaining examples into three parts as the independent part for each annotator. We measure inter-annotator agreement among the three annotators in the shared part using Fleiss’s kappa~\cite{Fleiss1971MeasuringNS}. The scores are 0.78 for correctness and 0.73 for completeness indicating substantial agreement among the three annotators. According to the evaluation results, we make the following observations.

\paragraph{Our approach can obtain high-quality step-by-step instructions with high correctness and completeness.} As shown in Figure~\ref{instruction_quality}, the correctness and completeness of the step-by-step instruction are very high, even though all the step-by-step instructions are generated by ChatGPT without any manual modification. Besides, the statistics of Table~\ref{tab:statistics} show that the average word count per step-by-step instruction is larger than the average word count per definition. These findings convince us that the detailed, correct, and complete step-by-step instruction can help language models to complete the target tasks more easily.  

\paragraph{Refining can improve the quality of the step-by-step instructions.}  Both correctness and completeness of the step-by-step instructions of the test tasks increase after refining. This observation indicates that it is possible to further refine the step-by-step instructions by progressively leveraging more relevant information about the target tasks via multiple interactions with ChatGPT. 

To deeply analyze the quality of the step-by-step instructions, we split the step-by-step instructions into four classes, which are correct and complete, correct but incomplete, complete but incorrect, and incorrect and incomplete. The percentages of them in the whole dataset including the training set and test set are 87.9\%, 2.8\%, 0.5\%, and 8.8\%. We show the representative examples in Appendix~\ref{sec:case}.

\subsection{Step-by-Step Instruction Tuning} \label{sec:tuning}
To incorporate the step-by-step instructions, we further tune the T5-LM model\footnote{It is obtained by further tuning T5 with the LM objective.}~\cite{raffel2020exploring} from the checkpoint of \modelName{}~\cite{supernaturalinstructions} on the training set. Besides the effective elements used by \modelName{} including the task definition and positive examples, we add the step-by-step instruction and combine it with the definition and positive examples. The negative examples and the explanations do not be utilized, as previous work~\cite{supernaturalinstructions} find they could hurt the model performance.  We train and test our models with the step-by-step instructions and call our method \textit{Step-by-Step Instruction Tuning}. The full instruction template we adopted is as follows. 

\begin{quote}
\textcolor{red}{Step by Step Instruction:} \textcolor{red}{\{\{instruction\_content\}\}}

Definition:\textcolor{blue}{\{\{task\_definition\}\}}

Positive Example 1$-$ 

Input:\textcolor{blue}{\{\{example1\_input\}\}}

Output:\textcolor{blue}{\{\{example1\_output\}\}}

Positive Example 2$-$  

···

Now complete the following example$-$  

Input:\textcolor{blue}{\{\{instance\_input\}\}}

Output:

\end{quote}

\{\{instruction\_content\}\} and \{\{instance\_input\}\} represent the step-by-step instruction and the input of the instance respectively. The full example will be fed to the T5-LM model to predict the output.

\section{Experiment}
\subsection{Dataset}
We evaluate our method on \nameShort{}~\cite{supernaturalinstructions} \footnote{Apache License 2.0}, which is a large benchmark of various NLP tasks and their natural language instructions. Following its split for evaluating English cross-task generalization, the training set consists of 757 different tasks for supervision and the test set contains 119 unseen tasks for evaluation. The  evaluation tasks are diverse and can be divided into 12 categories such as question answering, title generation, and cause-effect classification, covering both generation and classification. ROUGE-L~\cite{lin-2004-rouge} is adopted to evaluate the methods following~\citet{supernaturalinstructions} since it aligns well with human evaluation and can be easily applied to various tasks. To accurately estimate the performance, we also conduct human evaluation. 

\subsection{Training Details}
We conduct our experiments based on T5-LM. As for \modelName{}, we rerun the public code\footnote{https://github.com/yizhongw/Tk-Instruct} released by \citet{supernaturalinstructions} and report the results. To leverage the step-by-step instructions, we continue to train T5-LM models from the checkpoints of \modelName{} with a training epoch of 2. The learning rate is set to 1e-5 for the 11B model and 5e-5 for others. The batch size is set to 16 for the 3B and 11B models and 8 for others. The maximum input length of \modelName{} and our models is set to 1224, and the maximum output length is set to 128. Following the experimental setting adopted by \citet{supernaturalinstructions}, we use 100 instances per task for training and testing. More training details such as training time can be found in Appendix~\ref{training_details}.

\subsection{Baselines}
There are three kinds of baselines. One kind of models is the heuristic baselines. \textbf{Copying Instance Input}~\cite{supernaturalinstructions} copies the input of the test instance as the output. \textbf{Copying Demo Output}~\cite{supernaturalinstructions} randomly selects one input demonstration example and copies the output of the example as the predicted output. Another kind of models is large language models such as \textbf{T5-LM}~\cite{raffel2020exploring} and \textbf{GPT3}~\cite{brown2020language}. They take the input and directly generate the output. The remaining baselines are \textbf{T0}~\cite{wei2022finetuned}, \textbf{InstructGPT}~\cite{ouyang2022training}, and \textbf{\modelName{}}~\cite{supernaturalinstructions}. \textbf{T0} is trained on many natural language prompted datasets using multi-task learning and can generalize to unseen tasks. \textbf{InstructGPT} adopts the RLHF fine-tuning procedure to learn to follow instructions. \textbf{\modelName{}} is the most relevant baseline, which is built on T5-LM and trained using the instructions provided by~\citet{supernaturalinstructions}. We also use such instructions but we leverage the additional step-by-step instructions, which can provide the detailed intermediate steps for solving the tasks. More details about baselines are shown in Appendix~\ref{training_details}.

\begin{table}[tbp]
	\centering
        \resizebox{\linewidth}{!}{
	\begin{tabular}{l|l|l}
		\hline
		\multicolumn{1}{c|}{}  &\multicolumn{1}{l|}{\multirow{1}{*}{Methods}} & \multicolumn{1}{c}{ROUGE-L}  \\
		\hline
		\hline
		\multicolumn{1}{c|}{\multirow{2}{*}{\makecell{Heuristic\\ Baselines}}}  
		&\multicolumn{1}{l|}{Copying Instance Input}& \multicolumn{1}{l}{14.2}  \\ 
		&\multicolumn{1}{l|}{Copying Demo Output} & \multicolumn{1}{l}{28.5} \\
		\hline
		\multicolumn{1}{c|}{\multirow{2}{*}{\makecell{Language \\ Models}}}  &	\multicolumn{1}{l|}{T5-LM (11B)}	 &             \multicolumn{1}{l}{30.2}  \\ 
		&\multicolumn{1}{l|}{GPT3 (175B)} & \multicolumn{1}{l}{45.0} \\
		\hline
		
		\multicolumn{1}{c|}{\multirow{10}{*}{\makecell{Instruction-tuned \\ Models}}}
		& \multicolumn{1}{l|}{T0 (11B)}	 &             \multicolumn{1}{l}{32.3}  \\ 
  		&	\multicolumn{1}{l|}{InstructGPT (175B)}	 &             \multicolumn{1}{l}{52.1}  \\ 
            \cline{2-3}
		&	\multicolumn{1}{l|}{\modelName{} (Base)}	 &             \multicolumn{1}{l}{42.6}  \\ 
		&	\multicolumn{1}{l|}{+ \textit{Step-by-Step Instruction Tuning} }	 &             \multicolumn{1}{l}{43.6\textcolor{red}{($\uparrow$ 1.0)}}  \\ 
            \cline{2-3}
		&	\multicolumn{1}{l|}{\modelName{} (Large)}	 &             \multicolumn{1}{l}{48.0}  \\ 
		&	\multicolumn{1}{l|}{+ \textit{Step-by-Step Instruction Tuning} }	 &             \multicolumn{1}{l}{49.7\textcolor{red}{($\uparrow$ 1.7)}}  \\ 
            \cline{2-3}
		&	\multicolumn{1}{l|}{\modelName{} (3B)}	 &             \multicolumn{1}{l}{54.4}  \\ 
		&	\multicolumn{1}{l|}{+ \textit{Step-by-Step Instruction Tuning} }	 &             \multicolumn{1}{l}{56.3\textcolor{red}{($\uparrow$ 1.9)}}  \\ 
            \cline{2-3}
		&	\multicolumn{1}{l|}{\modelName{} (11B)}	 &             \multicolumn{1}{l}{60.0}  \\ 
		&	\multicolumn{1}{l|}{+ \textit{Step-by-Step Instruction Tuning} }	 &             \multicolumn{1}{l}{60.9\textcolor{red}{($\uparrow$ 0.9)}}  \\ 
		 \hline
   	
           \hline
        \end{tabular}
        }
    
  	\caption{Results on the unseen tasks in the test set of SUP-NATINST. The step-by-step instructions improve the cross-task generalization ability of baseline models with different model sizes (Base, Large, 3B, and 11B). }
	\label{tab:main_result}
\end{table}

\subsection{Main Results}
We show the results of \textit{Step-by-Step Instruction Tuning} in Table \ref{tab:main_result}. There are four key takeaways. First, incorporating the step-by-step instructions can improve the cross-task generalization of languages models. \textit{Step-by-Step Instruction Tuning} consistently improves \modelName{} across different model sizes (Base, Large, 3B, and 11B). This observation demonstrates the effectiveness of our approach, as the step-by-step instructions can provide the detailed and specific procedures for completing the target tasks while the original instructions are general. Second, overall performance increases with the model size, which can be concluded by comparing our models with different model sizes. This finding is in line with the scaling law. Third, instruction tuning can improve cross-task generalization of language models. For instance, InstructGPT, T0, and \modelName{} outperform their base models, namely GPT3, T5-LM, and T5-LM, as they have been trained to follow the instructions, which makes them benefit more from the instructions. The last one is that the heuristic baselines obtain worse results, which indicates that copying the input of the test instance or the output of the example without considering the task definition is far from enough.  Moreover, to estimate the model performance of ChatGPT, we randomly sample 10 instances for each test task resulting in 1190 instances for evaluation. ChatGPT obtains 61.2 ROUGE-L scores and our model (11B), which is much smaller than ChatGPT\footnote{https://openai.com/blog/chatgpt/}, surpasses ChatGPT and obtains 61.4 ROUGE-L scores on the same evaluation dataset. This result further reveals the effectiveness of our approach. 

\begin{table}[tbp]
	\centering
        \resizebox{\linewidth}{!}{
	\begin{tabular}{l|l}
		\hline
		\multicolumn{1}{l|}{\multirow{1}{*}{Methods}} & \multicolumn{1}{c}{ROUGE-L}  \\
		\hline
		\hline
		\multicolumn{1}{l|}{\modelName{} (3B)}	 &             \multicolumn{1}{l}{54.4}  \\ 
		\multicolumn{1}{l|}{+ \textit{Step-by-Step Instruction (Inference)} }	 & \multicolumn{1}{l}{55.0\textcolor{red}{($\uparrow$ 0.6)}}  \\ 
		\multicolumn{1}{l|}{+ \textit{Step-by-Step Instruction  (Training)} }	 &             \multicolumn{1}{l}{56.0\textcolor{red}{($\uparrow$ 1.6)}}  \\ 
		\multicolumn{1}{l|}{+ \textit{Step-by-Step Instruction  (Training \& Inference)} }	 &           \multicolumn{1}{l}{56.3\textcolor{red}{($\uparrow$ 1.9)}}  \\
		 \hline
        \end{tabular}
        }
  	\caption{Incorporating the step-by-step instructions in different phases.  }
	\label{tab:phrase}
\end{table}

\section{In-depth Analysis}
\subsection{Impact of Step-by-Step Instruction}
\paragraph{Incorporating the step-by-step instructions in either training or testing, or both phases helps cross-task generalization.} To analyze the effect of incorporating the step-by-step instructions in different phases, we conduct experiments based on \modelName{} (3B) and list the experimental results in Table~\ref{tab:phrase}. According to the results, we find that directly incorporating the step-by-step instructions in the inference phase can improve the model performance, even though the model does not see the step-by-step instructions in the training phase. Another observation is that leveraging the step-by-step instructions in the training phase can also boost the model performance and it is not necessary for obtaining improvement to fetch the step-by-step instructions during inference. Moreover, when we utilize the step-by-step instructions in both phases, our model achieves the best performance and surpasses the baseline model by 1.9 ROUGE-L scores. These findings demonstrate our motivation, which is that the step-by-step instructions can complete the original instructions.

\begin{table}[tbp]
	\centering
        \resizebox{\linewidth}{!}{
	\begin{tabular}{l|l|l}
		\hline
		\multicolumn{1}{l|}{\multirow{1}{*}{Methods}} & \multicolumn{1}{c}{Position} & \multicolumn{1}{c}{ROUGE-L}  \\
		\hline
		\hline
            \multicolumn{1}{l|}{\multirow{1}{*}{\modelName{} (Base)}}
		& \multicolumn{1}{l|}{None}	 &             \multicolumn{1}{l}{42.6}  \\ 
            \hline
            \multicolumn{1}{l|}{\multirow{2}{*}{+ \textit{Step-by-Step Instruction Tuning}}} & \multicolumn{1}{l|}{Prepend}	 &             \multicolumn{1}{l}{43.6\textcolor{red}{($\uparrow$ 1.0)}}  \\ 
            & \multicolumn{1}{l|}{Append}	 &             \multicolumn{1}{l}{43.6\textcolor{red}{($\uparrow$ 1.0)}}  \\ 
            \hline
            \hline
            \multicolumn{1}{l|}{\multirow{1}{*}{\modelName{} (Large)}}
		& \multicolumn{1}{l|}{None}	 &             \multicolumn{1}{l}{48.0}  \\ 
            \hline
            \multicolumn{1}{l|}{\multirow{2}{*}{+ \textit{Step-by-Step Instruction Tuning}}} & \multicolumn{1}{l|}{Prepend}	 &             \multicolumn{1}{l}{49.7\textcolor{red}{($\uparrow$ 1.7)}}  \\ 
            & \multicolumn{1}{l|}{Append}	 &             \multicolumn{1}{l}{50.3\textcolor{red}{($\uparrow$ 2.3)}}  \\ 
            \hline
            \hline
            \multicolumn{1}{l|}{\multirow{1}{*}{\modelName{} (3B)}}
		& \multicolumn{1}{l|}{None}	 &             \multicolumn{1}{l}{54.4}  \\ 
            \hline
            \multicolumn{1}{l|}{\multirow{2}{*}{+ \textit{Step-by-Step Instruction Tuning}}} & \multicolumn{1}{l|}{Prepend}	 &             \multicolumn{1}{l}{56.3\textcolor{red}{($\uparrow$ 1.9)}}  \\ 
            & \multicolumn{1}{l|}{Append}	 &             \multicolumn{1}{l}{55.3\textcolor{red}{($\uparrow$ 0.9)}}  \\ 

		 \hline
        \end{tabular}
        }
  	\caption{Incorporating the step-by-step instructions at different positions. ``Prepend'' and ``Append'' mean prepending and appending the step-by-step instructions to the task definitions respectively. }
	\label{tab:position}
\end{table}

\paragraph{Incorporating the step-by-step instructions at different positions.} To study the effect of different positions of the step-by-step instructions, we incorporate the step-by-step instructions by prepending and appending them to the task definitions. The experimental results are presented in Table~\ref{tab:position}. As shown in the table, for both positions, incorporating the step-by-step instructions can improve cross-task generalization across different model sizes, which demonstrates that the step-by-step instructions are useful for instructing the models to solve new tasks. According to the results, we find that prepending the step-by-step instructions to the task definitions is preferred considering the average improvement.

\begin{table}[tbp]
	\centering
        \resizebox{\linewidth}{!}{
	\begin{tabular}{l|l}
		\hline
		\multicolumn{1}{l|}{\multirow{1}{*}{Methods}} & \multicolumn{1}{c}{ROUGE-L}  \\
		\hline
		\hline
		\multicolumn{1}{l|}{\modelName{} (3B)}	 &             \multicolumn{1}{l}{54.4}  \\ 
            \hline
		\multicolumn{1}{l|}{+ \textit{Step-by-Step Instruction Inference} }	 &             \multicolumn{1}{l}{55.0}  \\
            \multicolumn{1}{l|}{+ \textit{Step-by-Step Instruction Inference (Original)} }	 & \multicolumn{1}{l}{54.8\textcolor{blue}{($\downarrow$ 0.2)}}  \\ 
            \hline
		\multicolumn{1}{l|}{+ \textit{Step-by-Step Instruction Tuning} }	 &             \multicolumn{1}{l}{56.3}  \\ 
  		\multicolumn{1}{l|}{+ \textit{Step-by-Step Instruction Tuning (Original)} }	 & \multicolumn{1}{l}{56.2\textcolor{blue}{($\downarrow$ 0.1)}}  \\ 
		 \hline
        \end{tabular}
        }
  	\caption{Results of ablating the refining process for obtaining the step-by-step instructions. \textit{Step-by-Step Instruction Inference} means only using the step-by-step instructions in the inference phase. }
	\label{tab:refined}
\end{table}

\subsection{Effect of the Refining Process}

As we mentioned in the previous section, the original step-by-step instructions generated by ChatGPT could not be good enough. Hence, we further ask ChatGPT to progressively refine the original step-by-step instructions. The human evaluation results shown in Figure~\ref{instruction_quality} have demonstrated that the correctness and completeness of the step-by-step instructions are improved through refining. To analyze the contribution of refining on the model, we conduct the ablation study and report the results in Table~\ref{tab:refined}. There are two findings. One is that refining the step-by-step instructions is helpful for improving the model performance on the unseen tasks since some mistakes could be fixed through refining leading to a better quality of the step-by-step instructions. The other one is that adopting the original step-by-step instructions can improve cross-task generalization, which further demonstrates the effectiveness of such instructions.

\begin{table}[tbp]
	\centering
        \resizebox{\linewidth}{!}{
	\begin{tabular}{l|l}
		\hline
		\multicolumn{1}{l|}{\multirow{1}{*}{Methods}} & \multicolumn{1}{c}{ROUGE-L}  \\
		\hline
		\hline
		\multicolumn{1}{l|}{\modelName{} (3B)}	 &             \multicolumn{1}{l}{54.4}  \\ 
            \hline
            \multicolumn{1}{l|}{+ \textit{Step-by-Step Instruction Inference} }	 &             \multicolumn{1}{l}{55.0}  \\
            \multicolumn{1}{l|}{+ \textit{Step-by-Step Instruction Inference (Shuffled)} }	 & \multicolumn{1}{l}{54.8\textcolor{blue}{($\downarrow$ 0.2)}}  \\ 
		
        \hline

		\multicolumn{1}{l|}{+ \textit{Step-by-Step Instruction Tuning} }	 &             \multicolumn{1}{l}{56.3}  \\ 
  		\multicolumn{1}{l|}{+ \textit{Step-by-Step Instruction Tuning (Shuffled)} }	 & \multicolumn{1}{l}{54.4\textcolor{blue}{($\downarrow$ 1.9)}}  \\ 
		 \hline
        \end{tabular}
        }
  	\caption{Results of randomly shuffling the order of steps of the step-by-step instruction.  }
	\label{tab:order}
\end{table}

\begin{table*}[tbp]
        \small
        \centering
        \resizebox{\linewidth}{!}{
	\begin{tabular}{p{0.3\linewidth} | p{0.7\linewidth}}
		\hline
		Task Category & Textual Entailment \\
		\hline
		Task ID & task190\_snli\_classification\\
		\hline
 		Task Definition & In this task, you're given a pair of sentences, sentence 1 and sentence 2. Your job is to choose whether the two sentences clearly agree (entailment)/disagree (contradiction) with each other, or if this cannot be determined (neutral). \textbf{Your answer must be in the form of the letters E, C, and N respectively.} \\
		\hline 
   		Step-by-Step Instruction & 1. Read sentence 1 and sentence 2. 2. Compare the two sentences to determine whether they agree or disagree with each other. 3. If the sentences agree with each other, choose the "entailment" option (E). 4. If the sentences disagree with each other, choose the "contradiction" option (C). 5. If it is not possible to determine whether the sentences agree or disagree, choose the "neutral" option (N).\\
		\hline 
            Instance Input & Sentence 1: Four males in a string quartet perform on an indoor stage. Sentence 2: The pianists put on shows in enormous outdoor arenas. \\
            \hline 
             Output (\modelName{})& E \quad \textcolor{blue}{\xmark} \\
             \hline
             Output (Ours)& C   \quad  \textcolor{red}{\cmark}\\
             \hline 

        \end{tabular}
        }

  	\caption{An example instance from task190\_snli\_classification in \nameShort{} benchmark.}
	\label{case_study}
\end{table*}

\subsection{Importance of the Order of Steps}
The order of the steps of the step-by-step instruction is vital for its correctness. For example, the step-by-step instruction for instructing the model to find the longest word in the given sentence is first splitting the sentence, obtaining the length of each word, and returning the longest word. If the order of the three steps is shuffled, the step-by-step instruction becomes: first obtaining the length of each word, splitting the sentence, and returning the longest word. As for this example, the step-by-step instruction is changed and is incorrect, which can hurt the effectiveness of the step-by-step instruction. But there are also some cases that shuffling the order does not hurt the correctness. For example, if we instruct the model to determine the relationship between the given two sentences,  there are two steps specifically first reading sentence A and then reading sentence B. Changing the order of these two steps does not affect the correctness.

To demonstrate this point, we automatically parse the step-by-step instructions and obtain the steps. Then, we randomly shuffle the steps for each step-by-step instruction. The experimental results are listed in Table~\ref{tab:order}. As we can see, randomly shuffling the step-by-step instructions hurts the performance especially for \textit{Step-by-Step Instruction Tuning}, as the noisy inputs make it hard to learn useful information through further tuning. 

\subsection{Human Evaluation}
We conduct the human evaluation to further demonstrate the effectiveness of the step-by-step instructions, as the automatic metric is a proxy of the performance of the models. Specifically, we randomly select three test instances for each task resulting in 357 instances. For each annotator, an instance is presented with the task definition, the associated positive samples, and two prediction results generated by our model (3B) and \modelName{} (3B). The annotators do not know where the predictions come from and determine which prediction is better. The win/lose/tie rates are 17.6\%, 12.6\%, and 69.8\%, and the Fleiss’s kappa score is 0.72,  which indicates our model achieves better performance and further demonstrates that \textit{Step-by-Step Instruction Tuning} can improve cross-task generalization.

\subsection{Case Study}
We show an example instance from the test set in Table \ref{case_study}. The task definition states that the answer must be in the form of the letters E, C, and N, but the meaning of these three options is not very clear only considering the last sentence. This requires language models should understand the context and find out that the letters E, C, and N represent entailment, contradiction, and neutral respectively. In contrast to it, the step-by-step instruction clearly explains the meaning of each option and provides relevant useful information about the three options in Step 3, Step 4, and Step 5. The step-by-step instruction elaborates the procedure of completing the textual entailment task, which enables our model to solve the problem step by step following it. With the help of such clear and detailed step-by-step instruction, our model (3B) completes the task successfully while \modelName{} (3B) fails.

\section{Discussion}

\paragraph{Standing on the shoulders of ChatGPT.} ChatGPT has impressed humans with its ability to provide detailed and well-organized answers to questions. Inspired by it, we propose to \textit{distill its knowledge of how to solve a specific problem} to improve the instruction and further enhance the cross-task generalization of language models via instruction tuning. Specifically, we treat ChatGPT as a meta-instructor and ask it to write down its step-by-step problem-solving process by prompting it. The smaller language model acts as an executor to follow the detailed instruction and complete the task.


\paragraph{Why can ChatGPT generate such useful step-by-step instructions? } ChatGPT is trained using instruction tuning and reinforcement learning from human feedback (RLHF), which enables it to understand the true intents of the human-written instructions and complete the tasks. Another possible reason is that pre-training on code data teaches it how to perform procedure-oriented programming and object-oriented programming, which helps it to learn to solve tasks step by step and decompose complex tasks into simpler ones~\cite{fu2022gptroadmap}.

\section{Conclusion}

 We propose to incorporate the step-by-step instructions to improve the cross-task generalization of language models. The extensive experiments demonstrate the effectiveness of our proposed approach, namely \textit{Step-by-Step Instruction Tuning}. We attribute the success to that the high-quality step-by-step instructions can provide the detailed and specific procedures for completing the tasks, which helps language models to decompose the tasks. We believe the step-by-step instructions can bring more opportunities to improve cross-task generalization. Hence, we release the step-by-step instructions and the corresponding results of the human evaluation for facilitating future research. For future work, we seek to enhance the quality of the step-by-step instructions including improving their correctness and completeness and making them more applicable for instructing language models.

\section*{Limitations}

As for limitations of our study, we obtain the step-by-step instructions by prompting ChatGPT and the biases of ChatGPT could affect the quality of the step-by-step instructions. Specifically, ChatGPT is tuned to interact with humans. So some generated detailed instructions are suitable for humans but not for the language models. For example, ChatGPT instructs to build a deep learning model for completing the sentiment classification task. To address this problem, one possible solution is manually refining the step-by-step instructions, but it is time-consumption and cost-consumption and can not be adopted quickly for new tasks. Another limitation is that we are not able to train larger models as we only can access the GPU resources. But we conduct extensive experiments and analyses with available language models such as T5-base, T5-Large, T5-3B, and T5-11B. The results demonstrate the effectiveness of our approach.  

\section*{Ethical Considerations}
The human evaluations were conducted by three well-educated students including two undergraduates and one graduate student. They were fully informed of the purpose of our study and the potential ethical risks involved in the tasks. We paid them at an hourly rate of \$7.37 USD per hour, which is a fair and reasonable hourly wage in our city. Specifically, each student was paid \$166.2 USD for evaluating the quality of the step-by-step instructions and \$10.8 USD for evaluating the prediction results. The collection of the step-by-step instructions through ChatGPT was compliant with the usage policies of OpenAI\footnote{https://beta.openai.com/docs/usage-policies}. We also manually checked all the generated step-by-step instructions to make sure the step-by-step instructions do not contain any offensive content or private information.

\bibliography{custom}
\bibliographystyle{acl_natbib}

\appendix

\section{Model Performance for Each Category} 
\label{sec:category}
\begin{figure*}[tbp]
\centering
\includegraphics[width=\linewidth]{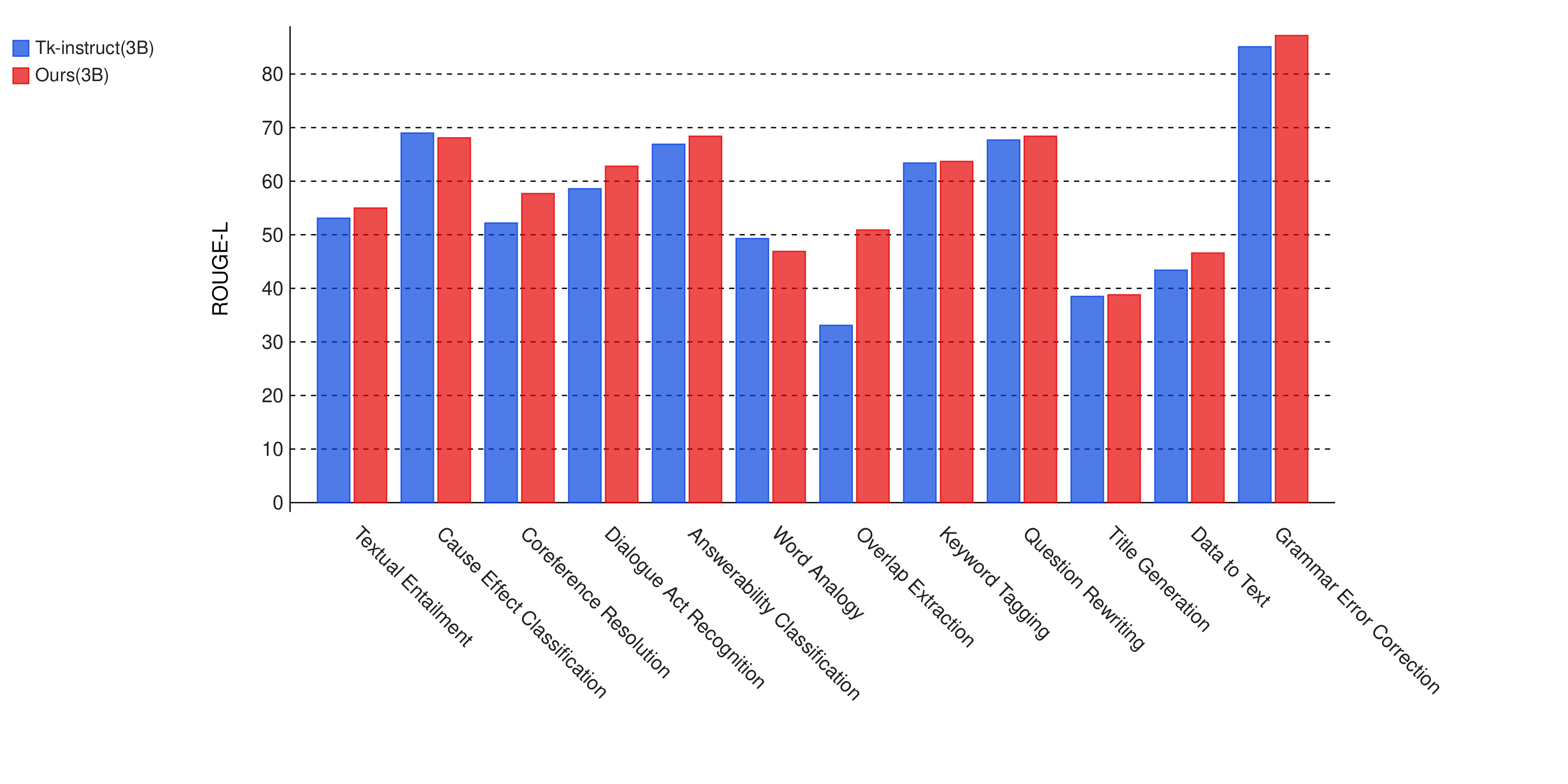} 
\caption{Model performance of our model (3B) and \modelName{} (3B) for each category. }
\label{improvement}
\end{figure*}

\begin{figure*}[tbp]
\centering
\includegraphics[width=\linewidth]{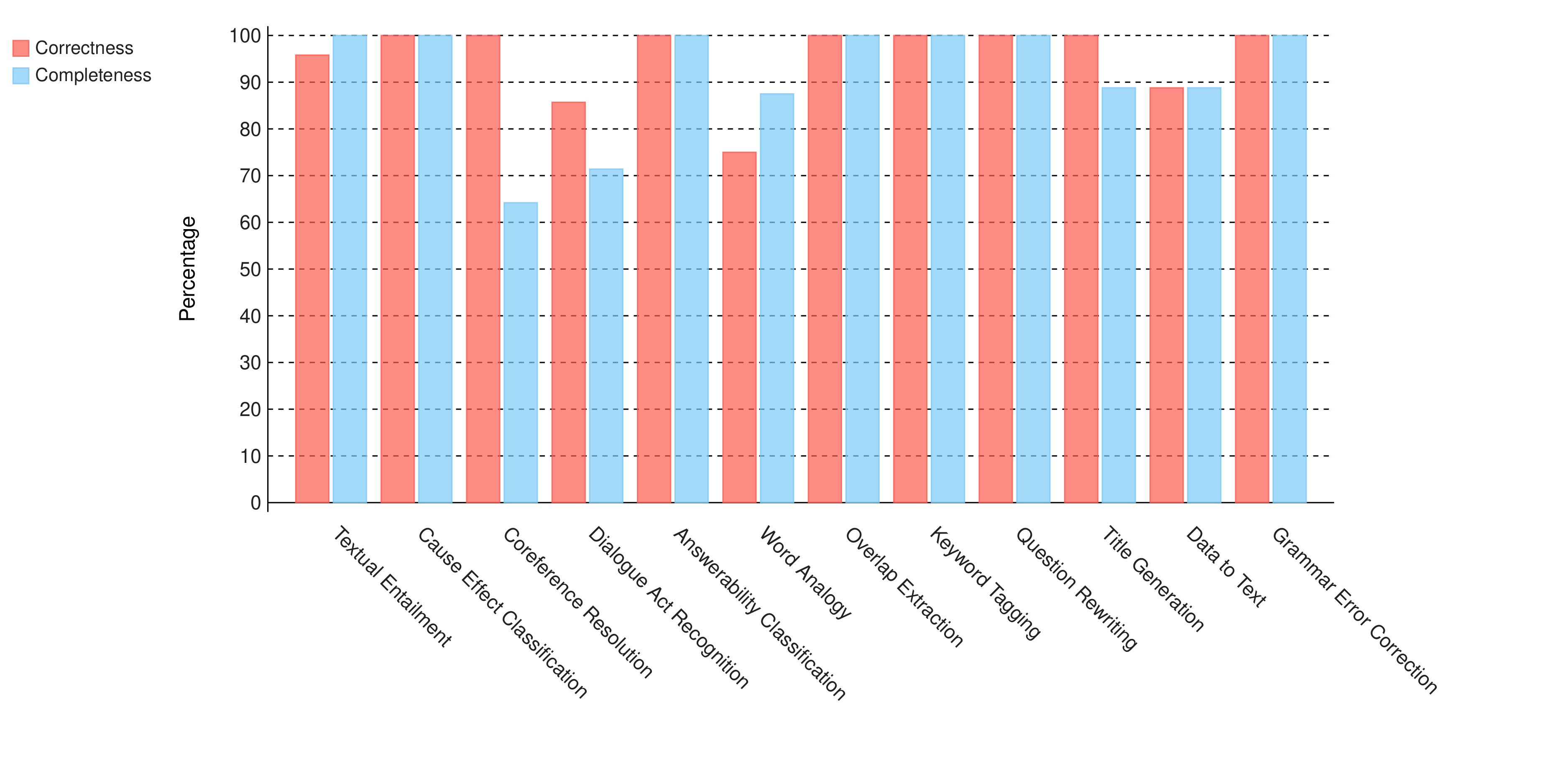} 
\caption{Correctness and completeness of the step-by-step instructions of the test dataset for each category. }
\label{test_quality}
\end{figure*}

\begin{figure*}[tbp]
\centering
\includegraphics[width=\linewidth]{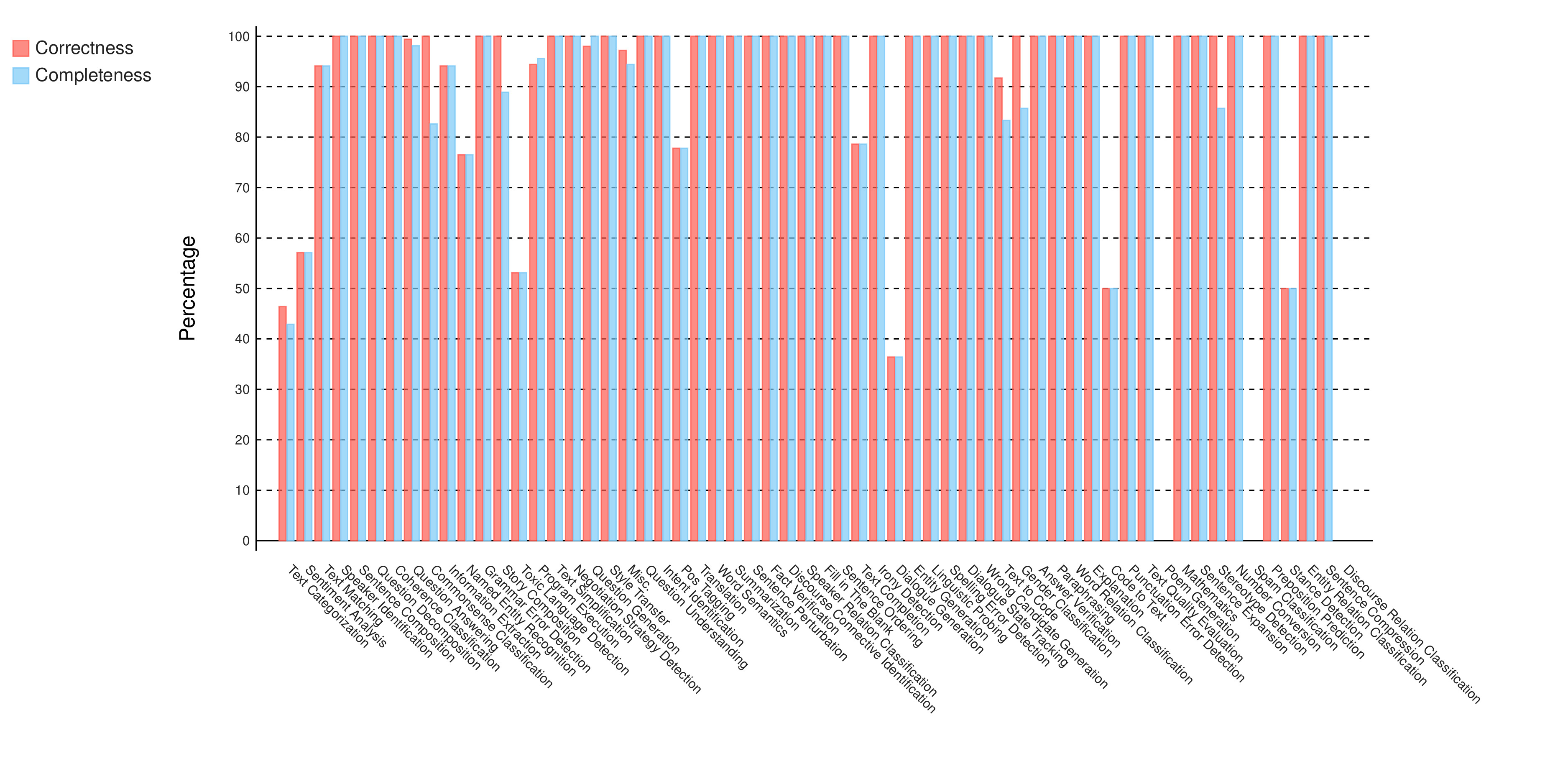} 
\caption{Correctness and completeness of the step-by-step instructions of the training dataset for each category. }
\label{train_quality}
\end{figure*}

We present the model performance of our model (3B) and \modelName{}(3B) for each category of the test dataset in Figure~\ref{improvement}. The results show that with \textit{Step-by-Step Instruction Tuning}, there is a gain in performance for most of the categories, especially for overlap extraction. One possible reason is that this kind of tasks could be solved more easily by following the steps of the step-by-step instructions, such as splitting the sentence, removing the stop words, and outputting the overlapping word. Another observation is that the model performance on word analogy drops and we attribute it to the low correctness of the step-by-step instructions as shown in Figure~\ref{test_quality}, as such instructions could provide wrong information and confuse the language model.  Besides, we also present the quality  of the step-by-step instructions of the training dataset in Figure~\ref{train_quality}. There are some categories such as spam classification, discourse relation classification, and poem generation, of which the correctness and completeness scores are 0. The reason is that there is only one task for these categories and the step-by-step instructions of them are labeled as incorrect and incomplete.

\section{Respective Examples}
\label{sec:case}
The respective examples are presented in Table~\ref{tab:correct_complete}, Table~\ref{tab:correct_incomplete}, Table~\ref{tab:incorrect_complete}, and Table~\ref{tab:incorrect_incomplete}. Table~\ref{tab:correct_complete} shows a correct and complete step-by-step instruction for sentiment analysis, which provides many details on how to finish the task. For example, it instructs to identify the emotion by paying attention to the sentiment words such as happy, sad, and angry. Table~\ref{tab:correct_incomplete} presents a correct but incomplete example, as the step-by-step instruction does not mention that the position of the pronoun is shown within two "\_"s, which is important information for completing the task. We consider the correct but incomplete examples also could bring benefits, as it does not warp the original meaning and could be complementary to the definition. In contrast to it, the incorrect examples shown in Table~\ref{tab:incorrect_complete} and Table~\ref{tab:incorrect_incomplete} could hurt the model performance. In the first example, the step-by-step instruction states ``Calculate the position of the lowercase alphabet in the English alphabet", but the listed examples are ``A is at position 1, B is at position 2", which is inconsistent with the previous instruction.  In the second example, the step-by-step instruction instructs to use natural language generation techniques to produce a response, which is applicable for language models. We attribute it to the bias of ChatGPT, which is tuned to produce human-desired results during training.

\begin{table}[tbp]
	\centering
        \resizebox{\linewidth}{!}{
	\begin{tabular}{l|l}
		\hline
		\multicolumn{1}{l|}{\multirow{1}{*}{Methods}} & \multicolumn{1}{c}{ROUGE-L}  \\
		\hline
		\hline
		\multicolumn{1}{l|}{\modelName{} (3B) \textit{w/ SSII \& w/o Examples} }	 &             \multicolumn{1}{l}{44.4}  \\ 
            \hline
            \multicolumn{1}{l|}{+ \textit{Generated Examples} }	 &             \multicolumn{1}{l}{51.6\textcolor{red}{($\uparrow$ 7.2)}}  \\
            \multicolumn{1}{l|}{+ \textit{Ranked Generated Examples} }	 & \multicolumn{1}{l}{51.7\textcolor{red}{($\uparrow$ 7.3)}}  \\ 
            \multicolumn{1}{l|}{+ \textit{Manually Written Examples} }	 & \multicolumn{1}{l}{55.0\textcolor{red}{($\uparrow$ 10.6)}}  \\         
            \hline
		\multicolumn{1}{l|}{\modelName{} (3B) \textit{w/ SSIT \& w/o Examples} }	 &             \multicolumn{1}{l}{44.0}  \\ 
            \hline
            \multicolumn{1}{l|}{+ \textit{Generated Examples} }	 &             \multicolumn{1}{l}{53.3\textcolor{red}{($\uparrow$ 9.3)}}  \\
            \multicolumn{1}{l|}{+ \textit{Ranked Generated Examples} }	 & \multicolumn{1}{l}{53.4\textcolor{red}{($\uparrow$ 9.4)}}  \\ 
            \multicolumn{1}{l|}{+ \textit{Manually Written Examples} }	 & \multicolumn{1}{l}{56.3\textcolor{red}{($\uparrow$ 12.3)}}  \\         
            \hline
            
        \end{tabular}
        }
  	\caption{Results of leveraging the generated demonstration examples by ChatGPT. SSII and SSIT mean \textit{Step-by-Step Instruction Inference} and \textit{Step-by-Step Instruction Tuning} respectively.  }
	\label{tab:examples}
\end{table}

\section{Quality of the Generated Examples}\label{generate_example}
We also attempt to ask ChatGPT to automatically generate the positive examples encouraged by the finding that it can generate valuable step-by-step instructions. To help ChatGPT to output desirable examples, we carefully design our prompts and introduce the self-ranking process. Specifically, we first ask ChatGPT to generate multiple examples. \{\{instruction\_content\}\} denotes the content of the step-by-step instruction and \{\{generated\_example\_num\}\} is a hyper-parameter which controls the number of the generated examples.

\begin{quote}
Give me \textcolor{red}{\{\{generated\_example\_num\}\}} harder examples for the \{\{task\_category\}\} task following this structure:

Input: 

Output: 

Explanation: 

The instruction for this task is :

\textcolor{red}{\{\{instruction\_content\}\}}
\end{quote}

Then let ChatGPT to rank the generated examples denoted as \{\{example\_content\}\} and return the Top-2 examples. 

\begin{quote}
Rank following examples by the correctness of their answers and the relevance and consistency with the task instruction. Return the content of the best two examples and do not need explain the reason.

\textcolor{red}{\{\{example\_content\}\}}

The instruction for this task is :
\textcolor{red}{\{\{instruction\_content\}\}}

\end{quote}

To qualify the quality of the obtained examples, we replace the manually written examples with the generated examples and the ranked examples. Note that, the number of all adopted examples is two for a fair comparison and the utilized generated examples are selected randomly from the obtained multiple examples. The experimental results are shown in Table~\ref{tab:examples}. According to the results, we find that generating the examples as better as the manually written examples is still very challenging since it requires the model to be able to fully understand the task and generate informative and right inputs and outputs. Even though the models utilizing the generated examples and ranked examples underperform the model adopting the manually written examples, they surpass the model without examples, which indicates that the generated and ranked examples are useful for improving cross-task generalization. Besides, the comparison between the models utilizing the generated examples and ranked examples shows the self-ranking process can further improve the quality of the obtained examples.

\section{More Experimental Details}\label{training_details}
\paragraph{Training.} We implement our approach based on the code released by \modelName{}\footnote{https://github.com/yizhongw/Tk-Instruct}. We train our 3B model on four A100 (40GB) GPUs and use DeepSpeed (0.7.7)\footnote{https://github.com/microsoft/DeepSpeed} to reduce the GPU memory. The training process takes 28 hours to complete. The 11B models are trained on four A100 (80GB) GPUs and it takes 120 hours to finish. As for the hyper-parameters, we mainly follow the hyper-parameters adopted by \modelName{} such as the learning rate and epoch number and we increase the max length of the input to 1224 for adding the step-by-step instruction. Note that, the results of \modelName{} reported in our paper are obtained by re-running \modelName{} with the increased length for a fair comparison. The random seed of all experiments in this paper is set to 42 following \modelName{}.

\paragraph{Evaluation.} Following the default setting of \nameShort{}~\cite{supernaturalinstructions}, we choose ROUGE-L as our metric. Specifically, the python package, namely rouge-score (0.1.2), is adopted, which is also used by \modelName{}~\cite{supernaturalinstructions}.  

\paragraph{GPT-3 and InstructGPT results.} We use the prediction results released by \modelName{}~\cite{supernaturalinstructions} to evaluate the performance of GPT-3 and InstructGPT. They accessed GPT-3 (``davinci'') and InstructGPT (``text-davinci-001'' ) through the API on May 30, 2022 and shared the prediction results\footnote{https://github.com/yizhongw/Tk-Instruct/tree/main/output/default}. We compute ROUGE-L of GPT-3 and InstructGPT based on the prediction results.

\section{Annotation Details}\label{annotation}
\paragraph{Annotators.} We invited three well-educated students including two undergraduates and one graduate student as our annotators to evaluate the quality of the step-by-step instructions and conduct the comparison between our model and the baseline model. We paid them a fair and reasonable hourly wage.

\paragraph{Instructions for Annotation.} (1) \textbf{Analysis of the quality of the step-by-step instruction}: you are given some examples and each example consists of a task definition, two positive examples, and a step-by-step instruction generated by ChatGPT. The step-by-step instruction will be combined with the task definition and the positive examples to instruct language models to complete the target task. Please carefully read the given example and analyze the correctness and completeness of the step-by-step instruction. As for correctness, if each step of the step-by-step instruction and the order of the steps are right for completing the target task, it is considered as correct.  As for completeness, if the step-by-step instruction contains all necessary information about the task definition such as the constraints, it is considered complete. (2) \textbf{Comparison between our model and the baseline model}: you are given some examples and each example consists of a task definition, two positive examples, and two predictions from two evaluated models (Model A and Model B). Please carefully read the given example and determine which prediction is better (``Model A'' or ``Model B''). If you can not determine which prediction to select, pick the ``Same'' option.

\paragraph{Intended Use.} The intended use of \nameShort{} ~\cite{supernaturalinstructions} is evaluating the generalization of language models to unseen tasks. Our work aims to improve generalization by incorporating the step-by-step instructions and we evaluate the effectiveness of our approach on \nameShort{} . In this paper, the use of \nameShort{} is consistent with its intended use. We also hope our obtained step-by-step instructions and the annotations could further facilitate research on improving cross-task generalization.

\begin{table*}[tbp]
	\begin{tabular}{p{0.3\linewidth} | p{0.7\linewidth}}
		\hline
		Task Category & Sentiment Analysis \\
		\hline
		Task ID & task293\_storycommonsense\_emotion\_text\_generation \\
		\hline
 		Task Definition & In this task, you're given a context, a sentence,  and a character. The sentence describes an action or job of the given character. Also, the context provides more information about the sentence or the character.  Your task is to return one of the emotions which are expressed by the Character in the given sentence. For that you can use the Context; however, Context is not provided in all the inputs. Also, in some inputs, there can be zero emotion; for that return "None".  \\
		\hline 
   		Step-by-Step Instruction & 1. Begin by reading the context and sentence provided in the input.  2. Identify the emotion or sentiment expressed by the character in the sentence. Look for words or phrases that indicate the character's emotional state, such as "happy," "sad," or "angry."  3. If the context is provided, consider how it may affect the character's emotional state. For example, if the context describes a positive event or experience, it may indicate that the character is feeling happy or content. 4. If you are unable to determine the character's emotion from the sentence or context, return "None". 5. Once you have identified the emotion expressed by the character, return that emotion as your output. \\
		\hline 
        \end{tabular}

  	\caption{A correct and complete step-by-step instruction for sentiment analysis.  }
	\label{tab:correct_complete}
\end{table*}

\begin{table*}[tbp]
	\begin{tabular}{p{0.3\linewidth} | p{0.7\linewidth}}
		\hline
		Task Category & Coreference Resolution \\
		\hline
		Task ID & task329\_gap\_classification \\
		\hline
 		Task Definition & In this task, you will be presented with a text, a pronoun from the text, and two candidate names. You should determine what the pronoun refers to and classify the answers into A, B, or Neither. A and B here are referring to option A and option B. \textbf{Position of the pronoun in the text is showed within two "\_"s.}  \\
		\hline 
   		Step-by-Step Instruction & 1. Read the given text and pronoun. 2. Identify the two candidate names, referred to as A and B. 3. Consider the context of the pronoun within the text, including the words and phrases surrounding it. 4. Determine which of the two candidate names, A or B, the pronoun most likely refers to based on the context. 5. Classify the answer as A, B, or Neither, depending on your determination. \\
		\hline 
        \end{tabular}

  	\caption{A correct but incomplete step-by-step instruction for question answering.  }
	\label{tab:correct_incomplete}
\end{table*}

\begin{table*}[tbp]
	\begin{tabular}{p{0.3\linewidth} | p{0.7\linewidth}}
		\hline
		Task Category & Program Execution \\
		\hline
		Task ID & task622\_replace\_alphabets\_in\_a\_list\_by\_their\_position\_in\_english\_alphabet\\
		\hline
 		Task Definition & In this task, you are given an input list A. You need to convert all the alphabets in the list with a number representing their position in the English alphabet. E.g., replace A by 1, B by 2, a by 1, b by 2, and so on.\\
		\hline 
   		Step-by-Step Instruction & 1. Initialize an empty output list, which will store the modified elements of the input list. 2. Iterate through each element in the input list. 3. For each element in the input list, check if it is an alphabet. If it is not an alphabet, append it to the output list without modifying it. 4. If the element is an alphabet, convert it to its lowercase form. 5. Calculate the position of the lowercase alphabet in the English alphabet \textbf{(e.g., A is at position 1, B is at position 2, etc.)}. 6. Append the calculated position to the output list. 7. Repeat the process for each element in the input list until all elements have been processed. 8. Return the output list as the result of the program execution.\\
		\hline 
        \end{tabular}

  	\caption{An incorrect but complete step-by-step instruction for program execution.  }
	\label{tab:incorrect_complete}
\end{table*}

\begin{table*}[tbp]
	\begin{tabular}{p{0.3\linewidth} | p{0.7\linewidth}}
		\hline
		Task Category & Dialogue Generation \\
		\hline
		Task ID & task1600\_smcalflow\_sentence\_generation\\
		\hline
 		Task Definition & In this task, you have given an input which is agent's reply, based on that you have to identify what will be users' command for that reply. The user generally asks a question related to event scheduling or information of the scheduled events. Agent reply based on the available information or do the task as per user command. \\
		\hline 
   		Step-by-Step Instruction & 1. Identify the input, which is the agent's reply to a user's question or command. 2. Determine the context of the conversation by considering the previous exchange between the user and the agent. 3. Use the context and the content of the agent's reply to generate a appropriate response or command from the user. 4. Consider the user's goals and needs, as well as any constraints or limitations, when generating the response. 5. \textbf{Use natural language generation techniques} to produce a response that is coherent and flows naturally with the rest of the conversation. 6. Test and iterate on the generated response to ensure it is accurate and appropriate for the given context.
 \\
		\hline 
        \end{tabular}

  	\caption{An incorrect and incomplete step-by-step instruction for dialogue generation.  }
	\label{tab:incorrect_incomplete}
\end{table*}

\end{document}